\documentclass{article}
\usepackage{spconf,amsmath,graphicx}

\usepackage{times}
\usepackage{epsfig}
\usepackage{graphicx}
\usepackage{amsmath}
\usepackage{amssymb}
\usepackage{dsfont}
\usepackage{amsfonts}
\usepackage{mathtools} 

\usepackage{amsthm}
\usepackage{booktabs}
\usepackage{tabularx}
\usepackage{algorithm}
\usepackage{algorithmic}
\usepackage[usenames,dvipsnames]{color}  
\usepackage{graphicx}    
\usepackage{url}
\usepackage{float}

\usepackage[update,prepend]{epstopdf}   
\graphicspath{figures/}
\usepackage{subcaption}

\def\relu{{\text{ReLU}}}

\title{End-to-end Binary Representation Learning via Direct Binary Embedding}
\name{Liu Liu, Alireza Rahimpour, Ali Taalimi, Hairong Qi}
\address{Department of Electrical Engineering and Computer Science\\
	University of Tennessee, Knoxville}
\begin{document}

\ninept
\clubpenalty=10000
\widowpenalty=10000

\setlength{\abovedisplayskip}{6pt}
\setlength{\belowdisplayskip}{6pt}
%
\maketitle
\begin{abstract}
Learning binary representation is essential to large-scale computer vision tasks. Most existing algorithms require a separate quantization constraint to learn effective hashing functions. In this work, we present Direct Binary Embedding (DBE), a simple yet very effective algorithm to learn binary representation in an end-to-end fashion. By appending an ingeniously designed DBE layer to the deep convolutional neural network (DCNN), DBE learns binary code directly from the continuous DBE layer activation without quantization error. By employing the deep residual network (ResNet) as DCNN component, DBE captures rich semantics from images. Furthermore, in the effort of handling multilabel images, we design a joint cross entropy loss that includes both softmax cross entropy and weighted binary cross entropy in consideration of the correlation and independence of labels, respectively. Extensive experiments demonstrate the significant superiority of DBE over state-of-the-art methods on tasks of natural object recognition, image retrieval and image annotation. 
\end{abstract}
\section{Introduction}
\label{sec:intro}

Representation learning is key to computer vision tasks. Recently with the explosion of data availability, it is crucial for the representation to be computationally efficient as well~\cite{Shen_2015_CVPR,cebits,alireza2017}. Consequently learning high-quality binary representation is tempting due to its compactness and representation capacity. 

Binary representation traditionally has been learned for image retrieval and similarity search purposes (image hashing). From the early works using hand-crafted visual features~\cite{5995432,spectralhashing,ksh,DBLP:conf/cvpr/LinSSHS14} to recent end-to-end approaches~\cite{aaai_ZhuL0C16,Liu_2016_CVPR,dsrh} that take advantages of deep convolutional neural networks (DCNN), the core of image hashing is learning binary code for images by characterizing the similarity in a pre-defined neighborhood. Usually pairwise or triplet similarity are considered to capture such similarity among image pairs or triplets, respectively~\cite{ksh,aaai_ZhuL0C16,Liu_2016_CVPR}. Albeit the high-quality of binary code, most image hashing algorithms do not consider learning discriminative binary representation. Recently this gap was filled by several hashing algorithms that learn binary representation via classification~\cite{dsrh,Shen_2015_CVPR,cebits}. Not only does the learned binary code retrieves images effectively, it provides comparable or even superior performance for classification as well. Meanwhile, due to the discrete nature of binary code, it is usually impractical to optimize discrete hashing function directly. Most hashing approaches attempt solving it by a continuous relaxation and quantization loss~\cite{Shen_2015_CVPR,Liu_2016_CVPR}. However, such optimization is usually not statistically stable~\cite{aaai_ZhuL0C16} and thus leads to suboptimal hash code.
 
In this work, we propose to learn high-quality binary representation \textit{directly} from deep convolutional neural networks (DCNNs). 
By appending a binary embedding layer directly into the state-of-the-art DCNN, deep residual network, we train the whole network as a hashing function via classification task in attempt to learning representation that approximates binary code without the need of using quantization error. Thus we name our approach Direct Binary Embedding (DBE). Furthermore, in order to learn high-quality binary representation for multilabel images, we propose a joint cross entropy that incorporates softmax cross entropy and weighted binary sigmoid cross entropy in consideration of the correlation and independence of labels, respectively. Extensive experiments on two large-scale datasets (CIFAR-10 and Microsoft COCO) show that the proposed DBE outperforms state-of-the-art hashing algorithms on object classification and retrieval tasks. Additionally, DBE provides a comparable performance on multilabel image annotation tasks where usually continuous representation is used.

\section{Direct Binary Embedding}
\label{sec: dbe}

\subsection{Direct Binary Embedding (DBE) Layer}
We start the discussion of DBE layer by revisiting learning binary representation using classification. Let $\mathbf{I} = \{I_i\}_{i=1}^N$ be the image set with $n$ samples, associated with label set $\mathbf{Y} = \{y_i\}_{i=1}^N$. We aim to learn binary representation $\mathbf{B}=\{\mathbf{b}_i\}_{i=1}^N\in\{0,+1\}^{N\times L}$ of $\mathbf{I}$ via the Direct Binary Embedding layer that is appended to DCNN. Following the paradigm of classification problem formulation in DCNN, we use a linear classifier $\mathbf{W}$ to classify the binary representation:
\begin{align}\label{eq:classification}
\min_{W,F}\ &\frac{1}{N}\sum_{i=1}^N\left(\mathcal{L}(\mathbf{W}^\top\mathbf{b}_i, y_i)+\lambda\|\mathbf{b}_i-F(I_i;\Omega)\|_2^2\right) \\
\text{s.t.}\ &\mathbf{b}_i = thresold(F(I_i;\Omega),0.5)\nonumber 
\end{align}
where $\mathcal{L}$ is an appropriate loss function; $\|\mathbf{b}_i-F(I_i;\Omega)\|_2^2$ measures the quantization error of between the DCNN activation $F(I_i;\Omega)$ and the binary code $\mathbf{b}_i$; $\lambda$ is the coefficient controlling the quantization error; $threshold(v,t)$ is a thresholding function at $t$, and it equals to $1$ if $v\geq t$, $0$ otherwise; $F$ is a composition of $n+1$ non-linear projection functions parameterized by $\Omega$:
\begin{equation}
F(I,\Omega) = f_{\text{DBE}}(f_n( \cdots f_2(f_1(I;\omega_1);\omega_2) \cdots;\omega_n)\omega_{\text{DBE}}),
\end{equation}
where the inner $n$ nonlinear projections composition denotes the $n$-layer DCNN; $f_{\text{DBE}}(\cdot;\omega_{\text{DBE}})$ is the Direct Binary Embedding layer appended to the DCNN. The binary code $\mathbf{b}_i$ in Eq.~\ref{eq:classification} makes it difficult to optimize via regular DCNN inference. We relax Eq.~\ref{eq:classification} to the following form where stochastic gradient descent is feasible:
\begin{align}\label{eq:relax}
\min_{W,F}&\dfrac{1}{N}\sum_{i=1}^N\left(\mathcal{L}(\mathbf{W}^\top F(I_i;\Omega),y_i)+\lambda ||2F(I_i;\Omega)-\mathbf{1}| - \mathbf{1}|^2 \right)
\end{align}
As proved by \cite{aaai_ZhuL0C16}, the quantization loss $||2F(I_i;\Omega)-\mathbf{1}| - \mathbf{1}|^2$ in Eq.~\ref{eq:relax} is an upper bound of that in Eq.~\ref{eq:classification}, making Eq.~\ref{eq:relax} an appropriate relaxation and much easier to optimize.

\begin{figure}[!h]
	\centering
    		  \includegraphics[width=3.0in]{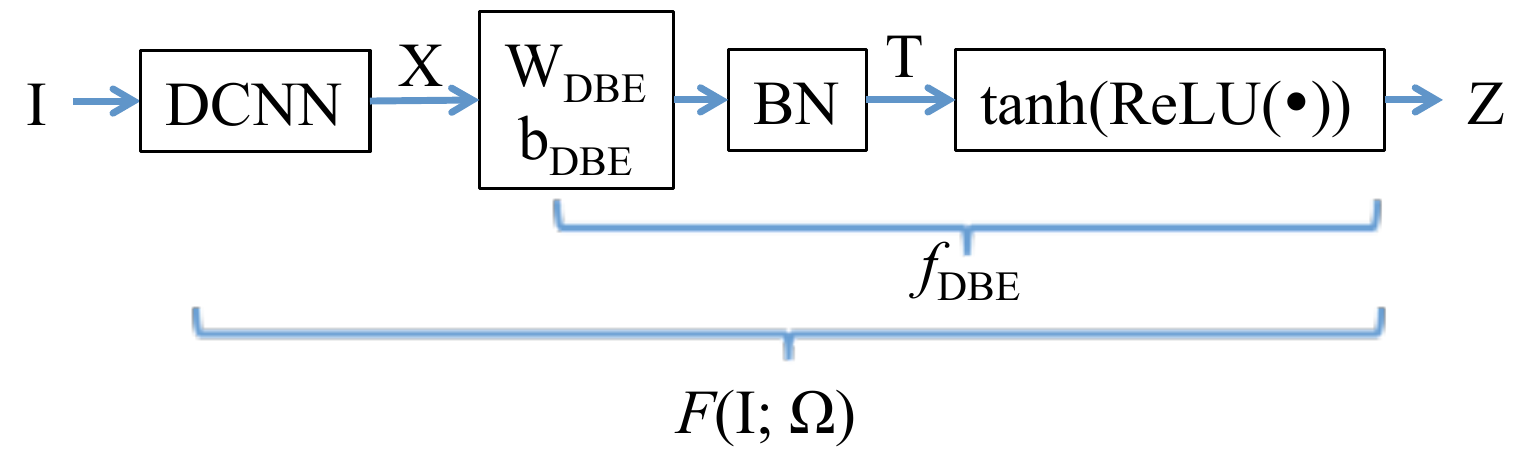}
  	\caption{The framework of DBE and outputs of different projections}
\label{fig:dbe_struc}
\end{figure}

\begin{figure}[h!]
    \centering
    \begin{subfigure}[h]{0.23\textwidth}
        \centering
        \includegraphics[height=1.0in]{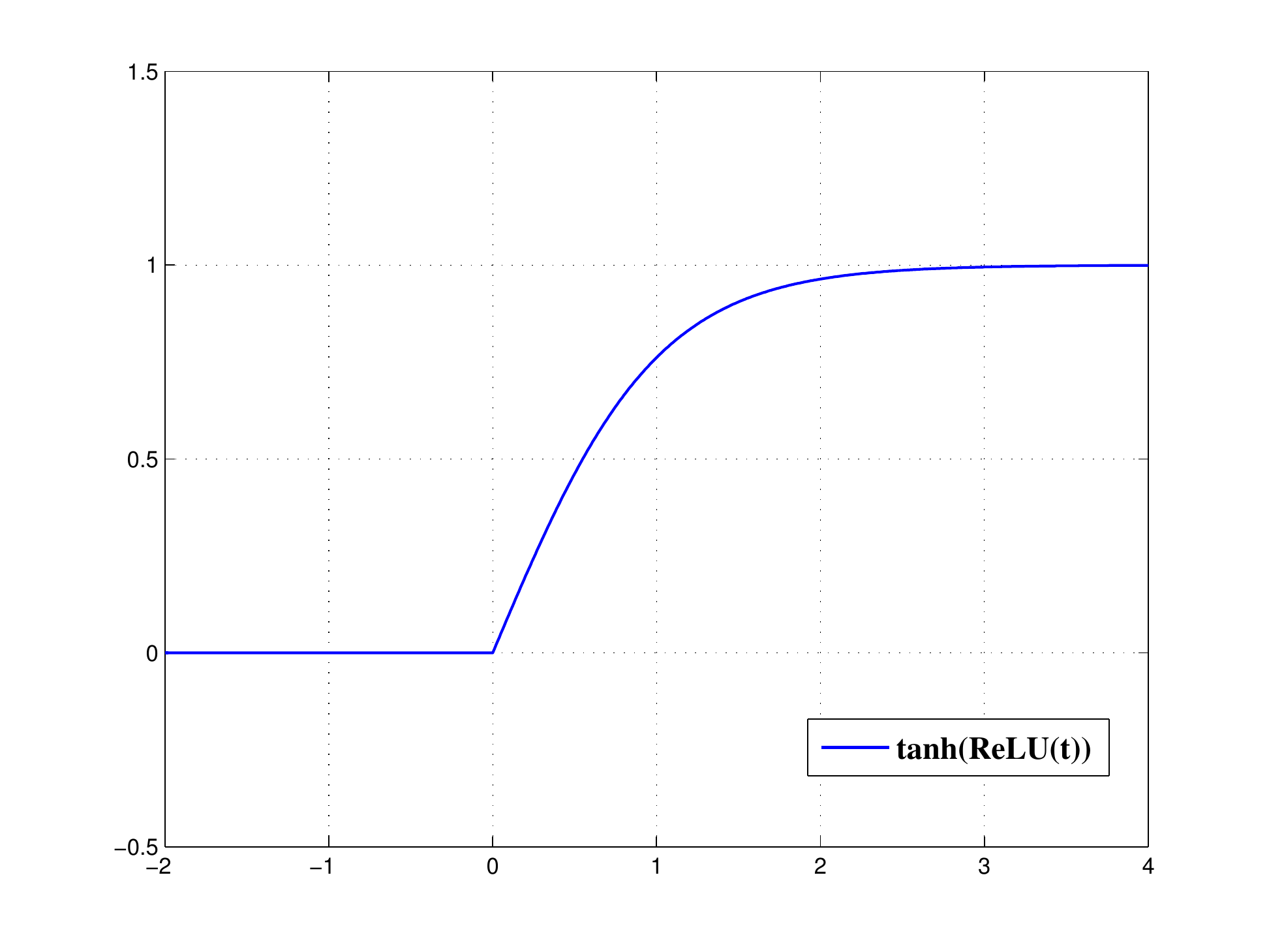}
        \caption{activation}
       	\label{fig:tanhrelu}
    \end{subfigure}%
    \begin{subfigure}[h]{0.23\textwidth}
        \centering
        \includegraphics[height=1.0in]{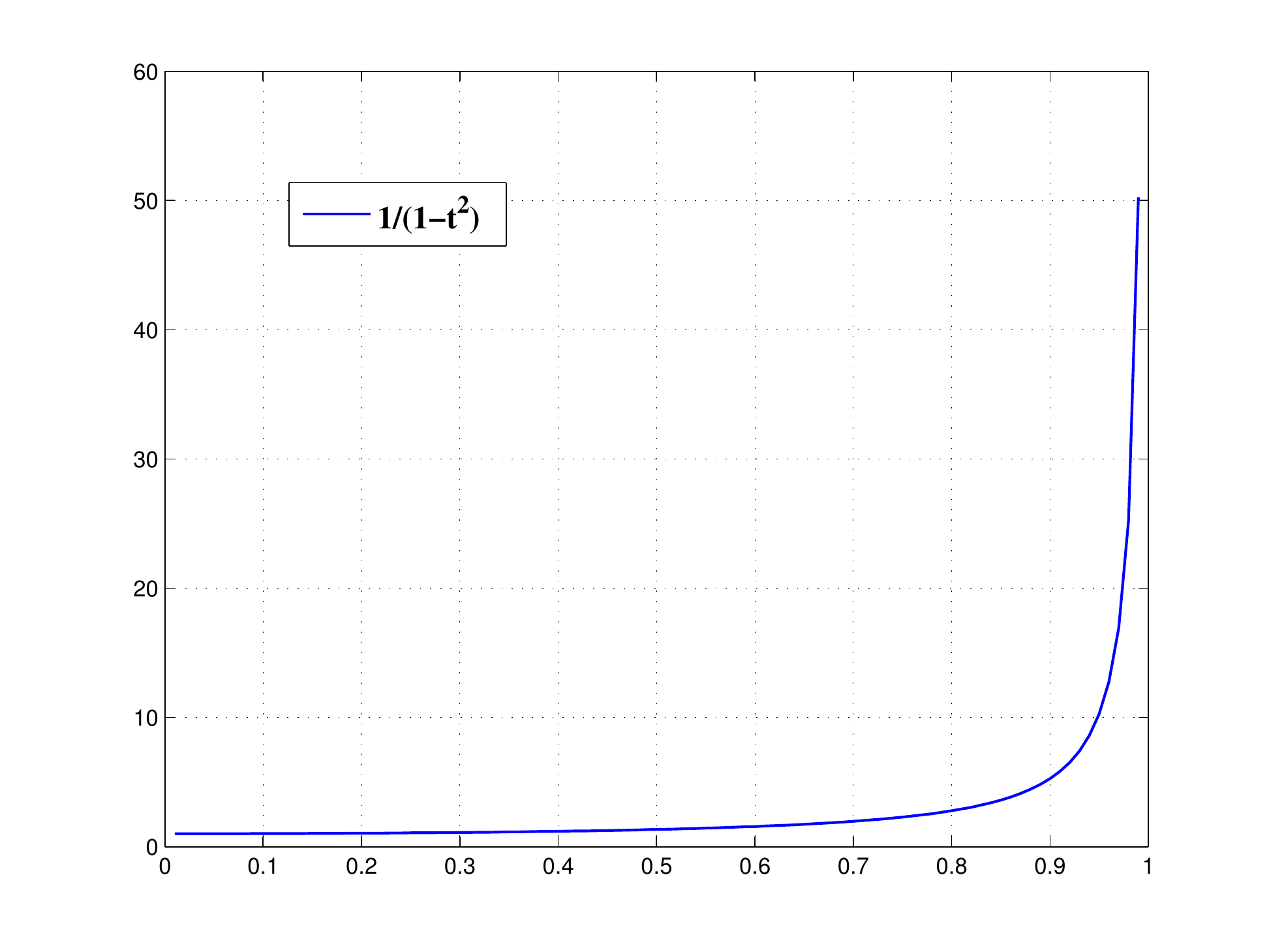}
        \caption{The PDF for positive input}
        \label{fig:pdf}
    \end{subfigure}
    \caption{$\tanh(\relu(\cdot))$ activation and its PDF for positive input}
\end{figure}

Several studies such as \cite{aaai_ZhuL0C16} share the similar idea of encouraging the fully-connected layer representation to be binary codes by using hyperbolic tangent (tanh) activation. Since it is desirable to learn binary code $\mathbf{B}=\{0,+1\}^{N\times L}$, we propose to concatenate the ReLU (rectified linear unit) nonlinearity with the tanh nonlinearity. Formally, we define DBE layer (shown in Figure~\ref{fig:dbe_struc}):
\begin{equation}\label{eq:dbe}
\mathbf{Z} = f_{\text{DBE}}(\mathbf{X}) = \tanh(\relu(\text{BN}(\mathbf{X}\mathbf{W}_{\text{DBE}}+b_\text{DBE})))
\end{equation}
where $\mathbf{X} = f_n( \cdots f_2(f_1(\mathbf{I};\omega_1);\omega_2) \cdots;\omega_n)\in \mathbb{R}^{N\times d}$ is the activation of $n$-layer DCNN; $\mathbf{Z} = f_{\text{DBE}}(\mathbf{X})\in \mathbb{R}^{N\times L}$ is the binary-like activation of DBE layer; $\mathbf{T}=\text{BN}(\mathbf{X}\mathbf{W}_{\text{DBE}}+b_\text{DBE})$ is the activation after linear projection and batch normalization but prior to ReLU and tanh; $\mathbf{W}_{\text{DBE}}\in\mathbb{R}^{d\times L}$ is a linear projection, $b_\text{DBE}$ is the bias; $\text{BN}(\cdot)$ is the batch normalization. And its activation is plotted in Figure~\ref{fig:tanhrelu}. The benefit of DBE layer approximating binary code is three-fold:
\begin{enumerate}
\itemsep0em 
\item batch normalization mitigates training with saturating nonlinearity such as tanh~\cite{bn}, and potentially promotes more effective binary representation.
\item ReLU activation is sparse~\cite{AISTATS2011_GlorotBB11} and learns bit `0' inherently.
\item tanh activation bounds the ramping of ReLU activation and learns bit `1' effectively without jeopardizing the sparsity of ReLU. 
\end{enumerate}
\vspace{-5mm}

Furthermore, DBE layer learns activation that approximates binary code statistically well. Consider random sampling  $t$ from $\mathbf{T}$, and assume it follows a distribution denoted by $p_T(t)$. Consequently the distribution of the DBE layer activation $z = f_{\text{DBE}}(t)$, and it follows distribution $p_Z(z)$, written as:
\begin{align}\label{eq:stats1}
&p_Z(z) = 
p_T(f_{\text{DBE}}^{-1}(z))\left|\frac{1}{f'_{\text{DBE}}(f_{\text{DBE}}^{-1}(z))}\right|
\end{align}
Eq.~\ref{eq:stats1} holds since $f_{\text{DBE}}$ is a monotonic and differentiable function. Since it is also positive when $z$ is positive, thus we have:
\begin{equation}\label{eq:dist_jacobian}
p_Z(z) = p_X(f_{\text{DBE}}^{-1}(z))\frac{1}{1-f_{\text{DBE}}^{-1}(z)^2}, \quad f_{\text{DBE}}^{-1}(z)=t>0.
\end{equation}
$p_T(f_{\text{DBE}}^{-1}(z))$ in Eq.~\ref{eq:dist_jacobian} is equivalent to $p_T(t)$; $\frac{1}{1-f_{\text{DBE}}^{-1}(z)^2}$ grows sharply towards the discrete value $\{+1\}$ for any positive response $z$, as is plotted in Figure~\ref{fig:pdf}. This suggests that the DBE layer enforces that the learned embedding $z$ are assigned to $\{+1\}$ with large probability as long as $z$ is positive. Conclusively DBE layer $f_{\text{DBE}}$ can effectively approximate binary code. Eventually we choose to optimize Eq.~\ref{eq:relax} without the quantization error and replace the binary code $\mathbf{b}_i$ with DBE layer activation directly. Eq.~\ref{eq:relax} can thus be rewritten as:
\begin{align}\label{eq:classification2}
\min_{\mathbf{W},F}\ &\frac{1}{N}\sum_{i=1}^N\mathcal{L}(\mathbf{W}^\top F(I_i;\Omega),y_i)\\ \nonumber
\text{s.t.}\ &F(\mathbf{I}, \Omega) = f_{\text{DBE}}(f_n( \cdots f_2(f_1(\mathbf{I};\omega_1);\omega_2) \cdots;\omega_n)\omega_{\text{DBE}})
\end{align}
The inference of DBE is the same as canonical DCNN models via stochastic gradient descent (SGD).

\subsection{Multiclass Image Classification}
\label{sec:multiclass}
Majority of DCNNs are trained via multiclass classification using softmax cross entropy as the loss function. Following this paradigm, Eq.~\ref{eq:classification2} can be instantiated as:
\begin{align}\label{eq:xen}
\min_{\mathbf{W},F}\ &-\frac{1}{N}\sum_{i=1}^N \sum_{k=1}^C\mathds{1}(y_i)\log\frac{e^{\mathbf{w}_k^\top F(I_i;\Omega)}}{\sum_{j=1}^C e^{\mathbf{w}_j^\top F(I_i;\Omega)}}\\
\text{s.t.}\ &F(\mathbf{I},\Omega) = f_{\text{DBE}}(f_n( \cdots f_2(f_1(\mathbf{I};\omega_1);\omega_2) \cdots;\omega_n)\omega_{\text{DBE}}) \nonumber
\end{align}
where $C$ is the number of categories; $\mathbf{W}=[\mathbf{w}_1, \ldots, \mathbf{w}_C]$ and $\mathbf{w}_k,\ k=1,\ldots,C$ is the weight of the classifier for category $k$; $y_i$ is the label for image sample $I$, and $\mathds{1}(y_i)$ an indicator function representing the probability distribution  for label $y_i$. Essentially Eq.~\ref{eq:xen} aims to minimize the difference between the probability distribution of ground truth label and prediction.

\subsection{Multilabel Image Classification}
\label{sec:multilabel}
More often a real-world image is associated with multiple objects belonging to different categories. A natural formulation of optimization problem for multilabel classification is extending the multiclass softmax cross entropy in Eq.~\ref{eq:xen} to multilabel cross entropy. Indeed softmax cross entropy captures the co-occurrence dependencies among labels, one cannot ignore the independence of each individual labels. For instance, `fork' and `spoon' usually co-exist in an image as they are associated with super-concept `dining'. But occasionally a `laptop' can be placed randomly on the dining table where there are also `fork' and `spoon' in the image as well. Consequently, we propose to optimize a joint cross entropy by incorporating weighted binary sigmoid cross entropy, which models each label independently, to softmax cross entropy. Eq.~\ref{eq:classification2} can therefore be instantiated as:
\begin{align}
\min_{\mathbf{W},F}\ &-\frac{1}{N}\sum_{i=1}^N \sum_{j=1}^{c_+}\frac{1}{c_+}\log\frac{e^{\mathbf{w}_j^\top F(I_i;\Omega)}}{\sum_{p=1}^C e^{\mathbf{w}_p^\top F(I_i;\Omega)}}\\ \nonumber
 &- \nu\frac{1}{N} \sum_{i=1}^N\sum_{p=1}^C \left[\rho\mathds{1}(y_i)\log\frac{1}{1+e^{\mathbf{w}_p^\top F(I_i;\Omega)}} \right.\\ \nonumber
 &\left. + (1-\mathds{1}(y_i))\log\frac{e^{\mathbf{w}_p^\top F(I_i;\Omega)}}{1+e^{\mathbf{w}_p^\top F(I_i;\Omega)}} \right]\\ 
\text{s.t.}\ &F(\mathbf{I};\Omega) = f_{\text{DBE}}(f_n( \cdots f_2(f_1(\mathbf{I};\omega_1);\omega_2) \cdots;\omega_n)\omega_{\text{DBE}}) \nonumber
\end{align}
where $c_+$ is the number of positive labels for each image; $\nu$ is the coefficient controlling the numerical balance between softmax cross entropy and binary sigmoid cross entropy; $\rho$ is the coefficient penalizing the loss for predicting positive labels incorrectly.

\subsection{Toy Example: LeNet with MNIST Dataset}
\begin{figure}[h!]
    \centering
    \begin{subfigure}[h]{0.23\textwidth}
        \centering
        \includegraphics[height=1.0in]{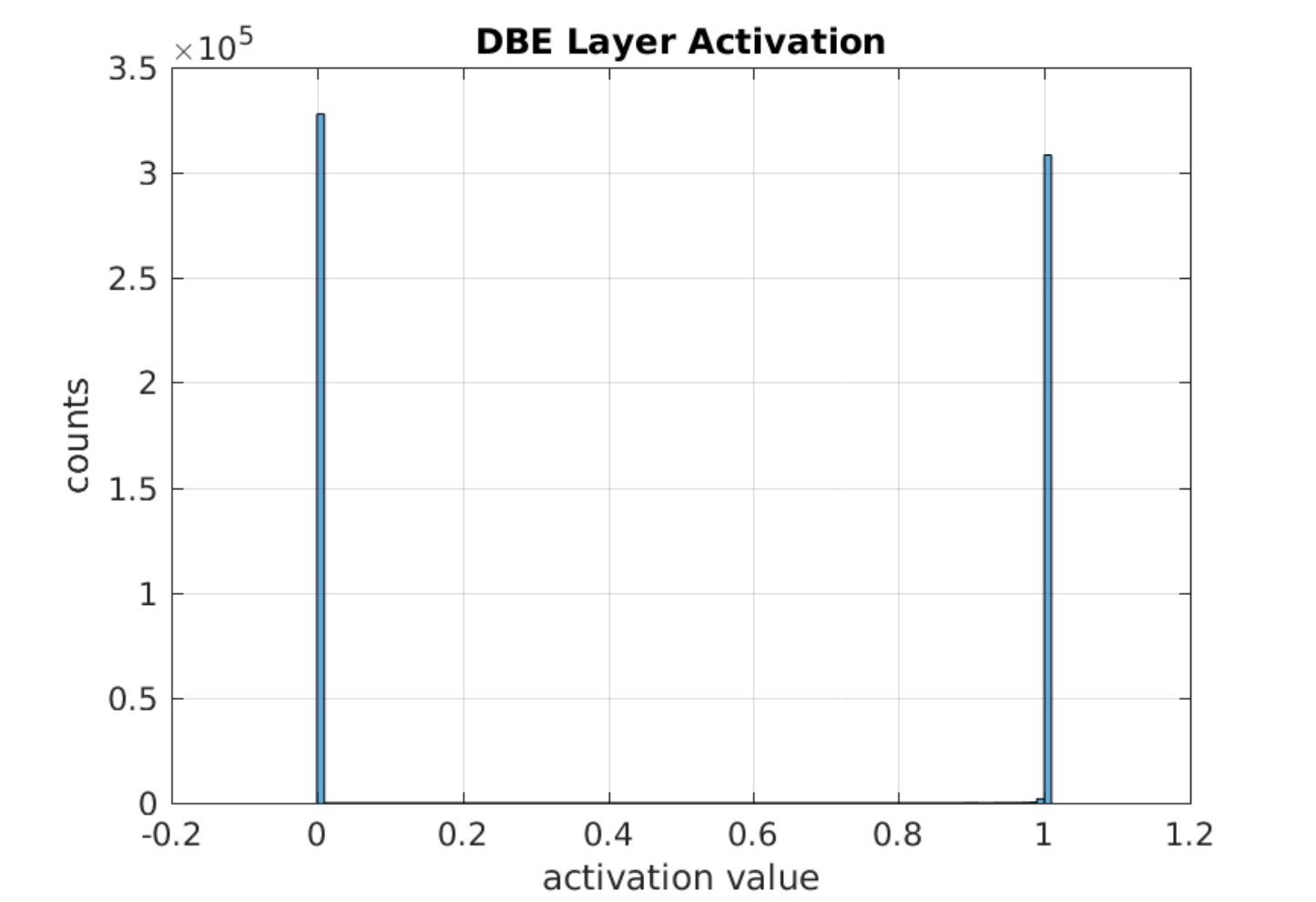}
        \caption{ }
       	\label{fig:dbe_act}
    \end{subfigure}%
    \begin{subfigure}[h]{0.23\textwidth}
        \centering
        \includegraphics[height=1.0in]{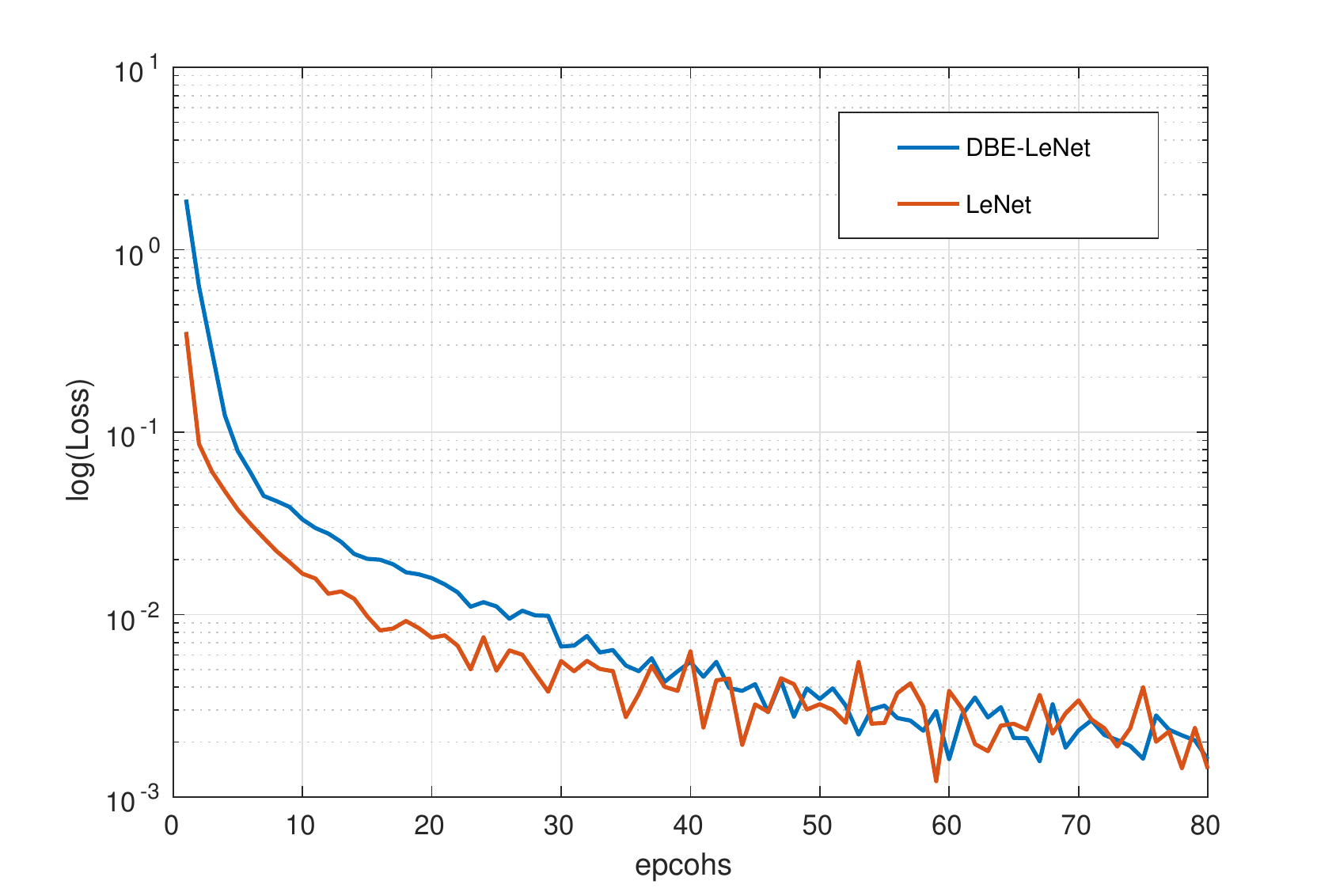}
        \caption{ }
        \label{fig:converg}
    \end{subfigure}
    \caption{The qualitatively results of DBE-LeNet: (a)The histogram of DBE layer activation; (b)The convergence of the original LeNet and with DBE trained on MNIST}
\end{figure}

In order to demonstrate the effectiveness of DBE layer, we use LeNet as a simple example of DCNN. We add DBE layer to the last fully connected layer of LeNet and learn binary representation for MNIST dataset. \textbf{MNIST} dataset~\cite{lecun-98} contains 70K hand-written digits of $28\times28$ pixel size, ranging from `0' to `9'. The dataset is split into a 60K training set (including a 5K validation set) and a 10K test set\footnote{http://yann.lecun.com/exdb/mnist/}. We enhance the original LeNet with more convolutional kernels (16 kernels and 32 kernels on the first and second layer, respectively, all with size $3\times 3$). We train the LeNet with DBE layer on the training set and evaluate the quality of learned binary representation on the test set. Figure~\ref{fig:dbe_act} demonstrates the histogram of activation from DBE. Clearly DBE layer learns a representation approximating binary code effectively (51.1\% of DBE activation less than 0.01, 48.6\% greater than 0.99 and only 0.3\% in between). We evaluate the quality of binary code learned by DBE qualitatively by comparing the classification accuracy on the test set with the state-of-the-art hashing algorithm. In order to demonstrate the effectiveness of DBE, we also compare with different $\lambda$ in Eq.~\ref{eq:relax} for the purpose of showing that quantization error is not necessary anymore to learn high-quality binary representation. From Table~\ref{tab:lambda} we can see that with the increase of $\lambda$ in Eq.~\ref{eq:relax}, the testing accuracy decreases. Due to the effectiveness of DBE layer, quantization error does not contribute to the binary code learning. Following the evaluation protocol of previous works~\cite{Shen_2015_CVPR}, linear-SVM~\cite{REF08a} is used as the classifier on all compared methods for fair comparison (including continuous LeNet representation). The classification accuracy on the test set is reported in Table~\ref{tab:mnist_acc}.

\vspace{-5mm}
\begin{table}[!htbp]
\centering
\scalebox{0.9}{
\begin{tabular}{ *5c }
	\toprule
	Method & LeNet~\cite{lecun-98} & DBE-LeNet	&	SDH~\cite{Shen_2015_CVPR} & FastHash~\cite{DBLP:conf/cvpr/LinSSHS14}	\\
	\midrule
	testing acc(\%)	&	99.34	&99.34	&	99.14 & 98.62	\\
	\bottomrule
\end{tabular}
}
\caption{The comparison of the testing accuracy on MNIST. Code-length for all hashing algorithms is 64-bit. LeNet feature (1000-d continuous vectors) is used for SDH and FastHash.}
\label{tab:mnist_acc}
\end{table}
\vspace{-5mm}
\begin{table}[!htbp]
\centering
\scalebox{0.9}{
\begin{tabular}{ *6c }
	\toprule
	$\lambda$ 	& 0		&1e-4	&1e-3	&1e-2	&1e-1\\
	\midrule
	testing acc(\%)	&	99.34	&99.34	&	99.30 & 99.26	&99.01	\\
	\bottomrule
\end{tabular}
}
\caption{The impact on quantization error coefficient $\lambda$}
\label{tab:lambda}
\end{table}


The convergence of training DBE-LeNet is reported in Figure~\ref{fig:converg}. Due to the saturating tanh activation, the gradient is slightly more difficult to propagate through the network. Eventually the convergence reaches the same level.


\section{Experiments}
We evaluate the proposed DBE layer with the deep residual network (ResNet).
We choose to append DBE layer to the state-of-the-art DCNN, 50-layer Residual Network (ResNet-50)~\cite{He_2016_CVPR} to learn high-quality binary representation for image sets. For the multilabel experiments, we set $\nu=2$ and $\rho=5$ through extensive empirical study.

\subsection{Dataset}
\textbf{CIFAR-10} dataset~\cite{Krizhevsky09learningmultiple} contains 60K color images (size $32\times32$) with each image containing a natural object. There are 10 categories of objects in total, with each category containing 6K images. The dataset is randomly split into a 50K training set and a 10K testing set. For traditional image hashing algorithms, we provide 512-D GIST~\cite{gist} feature; for end-to-end deep hashing algorithms, we use raw images as input directly.
\textbf{Microsoft COCO 2014 (COCO)}~\cite{Lin2014} is a dataset for image recognition, segmentation and captioning. It contains a training set of 83K images with 605K annotations and a validation set of 40K images with 292K annotations. There are totally 80 categories of annotations. We treat annotations as labels for images. On average each image contains 7.3 labels. Since images in COCO are color images with various sizes, we resize them to $224\times 224$.

\subsection{Object Classification}\label{sec:multiclass}
To evaluate the capability of mulitclass object classification, we compare DBE with several state-of-the-art supervised approaches including FastHash~\cite{DBLP:conf/cvpr/LinSSHS14}, SDH~\cite{Shen_2015_CVPR}, CCA-ITQ~\cite{5995432} and deep method DLBHC~\cite{Lin_2015_CVPR_Workshops}. The ResNet-50 features are also included in the comparison. The code-length of binary code from all the hashing methods is 64 bits. We use linear-SVM to evaluate the all the approaches on the classification task.

Table~\ref{tab:cifar_acc} shows the classification accuracy on test sets for the two datasets. The accuracy achieved by DBE matches that of the original continuous ResNet-50 features. DBE improves the state-of-the-art traditional methods and end-to-end approaches by 28.6\% and 5.6\%, respectively. And it achieves the same performance as that of the original ResNet. This demonstrates 1) DBE's superior capability of preserving the rich semantic information extracted by ResNet, 2) there exists great redundancy in the original ResNet features.
\begin{table}[!htbp]
\centering
\begin{tabular}{ l  c }
	\toprule
	Methods 	& 	Testing Accuracy (\%)  	\\ 
	\midrule
	CCA-ITQ~\cite{5995432}		&56.34 \\
	FastHash~\cite{DBLP:conf/cvpr/LinSSHS14}		&	57.82\\
	SDH~\cite{Shen_2015_CVPR}			& 67.73 \\
	DLBHC~\cite{Lin_2015_CVPR_Workshops}	& 86.73\\
	ResNet~\cite{He_2016_CVPR}	&\textbf{92.38}\\	
	DBE (ours) 				&\underline{92.35}\\
	\bottomrule
\end{tabular}
\caption{The testing accuracy of different methods on CIFAR-10 dataset. All binary representations have code-length of 64 bits.}
\label{tab:cifar_acc}
\end{table}

Furthermore we also provide the classification accuracy on CIFAR-10 with respect to different code lengths in Table~\ref{tab:cifar10_length}. From the table we can conclude that DBE learns high-quality binary representation consistently.
\begin{table}[!htbp]
\centering
\scalebox{1.0}{
\begin{tabular}{ *6c }
	\toprule
	Code length (bits) 		& 16 	& 	32	&	48 	& 64		& 128 \\
	\midrule
	testing acc(\%)	&91.63 	&	92.04	& 92.20 	& 92.35	& 92.36 \\
	\bottomrule
\end{tabular}
}
\caption{Classification accuracy of DBE on CIFAR-10 dataset across different code lengths}
\label{tab:cifar10_length}
\end{table}

\subsection{Image Retrieval}
\subsubsection{Natural Object Retrieval}
The CIFAR-10 dataset is used to evaluate the proposed DBE on natural object retrieval task. We choose to compare with state-of-the-art image hashing algorithms including both traditional hashing methods: CCA-ITQ~\cite{5995432}, FastHash~\cite{DBLP:conf/cvpr/LinSSHS14}, and end-to-end deep hashing methods: DSH~\cite{7780596}, DSRH~\cite{dsrh}. For the experimental settings, we randomly select 100 images per category and obtain a query set with 1K images. Mean average precision (mAP) is used as the evaluation metric. The comparison is reported in Table~\ref{tab:cifar10_map}. The proposed DBE outperforms state-of-the-art by around 3\%. It confirms that DBE is capable of preserving rich semantics extracted by the ResNet from original images and learning high-quality binary code for retrieval purpose.
\begin{table}[!htbp]
\centering
\scalebox{1.0}{
\begin{tabular}{ l c c c c }
	\toprule
	Code length (bits) 		& 12 	& 	24	&	36 	& 48	 \\
	\midrule
	CCA-ITQ~\cite{5995432}	&0.261	&0.289	&0.307	&0.310\\
	FastHash~\cite{DBLP:conf/cvpr/LinSSHS14}&0.286	&0.324	&0.371	&0.382\\
	SDH~\cite{Shen_2015_CVPR}&0.342	&0.397	&	0.411	&0.435\\	
	DSH~\cite{7780596}		&0.616	&0.651	&0.661	&	0.676\\
	DSRH~\cite{dsrh}			&0.792	&0.794	&0.792	&0.792\\
	DLBHC~\cite{Lin_2015_CVPR_Workshops}	&0.892	&0.895	&0.897	&0.897\\
	DBE (ours)		&\textbf{0.912}	&\textbf{0.924}	&\textbf{0.926}	&\textbf{0.927}\\
	\bottomrule
\end{tabular}
}
\caption{Comparison of mean average precision (mAP) on CIFAR-10}
\label{tab:cifar10_map}
\end{table}

\vspace{-5mm}
\subsubsection{Multilabel Image Retrieval}
COCO dataset is used for multilabel image retrieval task. Considering the large number of labels in COCO, we compare DBE with several cross modal hashing and quantization algorithms. Studies have shown that cross-modal hashing improves unimodal methods by leveraging semantic information of text/label modality~\cite{rastegari:icml13,Irie_2015_ICCV}. We choose to compare with CMFH~\cite{Ding_2014_CVPR} and CCA-ACQ~\cite{Irie_2015_ICCV}. Furthermore we also include traditional hashing method CCA-ITQ~\cite{5995432} and end-to-end approach DHN~\cite{aaai_ZhuL0C16}. Following the experiment protocols in \cite{Irie_2015_ICCV}, 1000 images are randomly sampled from validation set for query and the training set is used for database for retrieval. And AlexNet~\cite{DBLP:conf/nips/KrizhevskySH12} feature is used as input for algorithms that are not end-to-end, and raw images are used for end-to-end deep hashing algorithms. Due to the multilabel nature of COCO, we consider the true neighbors of a query image as the retrieved images sharing at least one labels with the query. Similar to natural object retrieval, mean average precision (mAP) is used as evaluation metric.
\begin{table}[!htbp]
\centering
\scalebox{0.9}{
\begin{tabular}{ l c c c c c}
	\toprule
	Code length (bits) 		& 16 	& 	24	&	32 	& 48		&	64	 \\
	\midrule
	CCA-ITQ~\cite{5995432}	&0.477	&0.481	&0.485	&0.490	&	0.494	\\
	CMFH	~\cite{Ding_2014_CVPR}		&0.462	&0.476	&0.484	&0.497	&0.505		\\
	CCA-ACQ~\cite{Irie_2015_ICCV}	&0.483	&0.500	&0.504	&0.515	&0.520	\\
	DHN~\cite{aaai_ZhuL0C16}	&0.507 	&	0.539	& 0.550 	& 0.559 &0.570\\
	DBE (ours)		&\textbf{0.623}	&\textbf{0.657}	&\textbf{0.670}	&\textbf{0.692}&\textbf{0.716}\\
	\bottomrule
\end{tabular}
}
\caption{Comparison of mean average precision (mAP) on COCO }
\label{tab:mscoco_map}
\end{table}

\subsection{Multilabel Image Annotation}
We generate prediction of labels for each image in validation set based on $K$ highest ranked labels and compare to the ground truth labels. The overall precision (O-P), recall (O-C), and F1-score (O-F1) of the prediction are used as evaluation metrics. Formally they are defined as:
\begin{align}
\text{O-P} = \frac{N_{CP}}{N_P},\ \text{O-R} = \frac{N_{CP}}{N_G},\ \text{O-F1} = 2\frac{\text{O-P}\cdot \text{O-R}}{\text{O-P}+\text{O-R}}
\end{align}
where $C$ is the number of annotations/labels; $N_{CP}$ is the number of correctly predicted labels for validation set; $N_P$ is the total number of predicted labels; $N_G$ is the total number of ground truth labels for validation set.

We compare DBE with softmax, binary cross entropy and WARP~\cite{gongJLTI14}, one of the state-of-the-art for multilabel image annotation. The performance comparison is summarized in Table~\ref{tab:coco_pred} and we set $K=3$ in the experiment. It can be observed that the binary representation learned by DBE achieves the best performance in terms of overall-F1 score. Due to its consideration of co-occurrence and independence of labels, DBE-joint cross entropy outperforms DBE-softmax and DBE-weighted binary cross entropy.
\begin{table}[!htbp]
\centering
\scalebox{0.9}{
\begin{tabular}{ l  c c c  }
	\toprule
	Method 							& O-P 	& 		O-R		&	O-F1  \\
	\midrule
	WARP~\cite{gongJLTI14}			&\textbf{59.8} 	& 61.4			& 	60.6 \\
	DBE-Softmax						&59.1	& 62.1		&	60.3	 \\
	DBE-weighted binary cross entropy	&57.1	& 60.8			&	58.9 \\
	DBE-joint cross entropy			& 59.5	& \textbf{62.7}			&	\textbf{61.1}	\\
	\bottomrule
\end{tabular}
}
\caption{Performance comparison on COCO for $K=3$. The code length for all the DBE methods is 64-bit.}
\label{tab:coco_pred}
\end{table}

\vspace{-5mm}
\subsection{The Impact of DCNN Structure}
Similar to most deep hashing algorithms, DBE also preserves semantics from DCNN. Consequently the structure of DCNNs influences the quality of binary code significantly. We compare with the state-of-the-art DLBHC~\cite{Lin_2015_CVPR_Workshops} and the DCNN it uses: AlexNet~\cite{DBLP:conf/nips/KrizhevskySH12}, which the upper bound in this comparison. Since DLBHC uses AlexNet, we also use AlexNet in our DBE. CIFAR-10 dataset is used. According to results reported in Table~\ref{tab:cifar_binarycompare}, DBE achieves higher accuracy than DLBHC, i.e., DBE learns more semantic and discriminative binary representation.
\begin{table}[!htbp]
\centering
\scalebox{0.9}{
\begin{tabular}{ *4c }
	\toprule
	Method 	& AlexNet~\cite{DBLP:conf/nips/KrizhevskySH12}	&  DBE-AlexNet& DLBHC~\cite{Lin_2015_CVPR_Workshops}		\\
	\midrule
	testing acc(\%)		&	89.20 (upper bound) &88.52& 86.73	\\
	\bottomrule
\end{tabular}
}
\caption{The comparison of the classification accuracy on the test set of CIFAR-10. Code-length for all binary algorithms is 48-bit.}
\label{tab:cifar_binarycompare}
\end{table}

\vspace{-5mm}
\section{Conclusion}
\label{sec:concl}

We proposed a novel approach to learn binary representation for images in an end-to-end fashion. By using a Direct Binary Embedding layer, we are able to approximate binary code directly in DCNN. Different from existing works, DBE learns high quality binary representation for images without quantization error as a regularization. Extensive experiments on two large-scale datasets demonstrate the effectiveness superiority of DBE over state-of-the-art on several computer vision tasks including object recognition, image retrieval and multilabel image annotation.

\bibliographystyle{IEEEbib}
\bibliography{icip17}

\end{document}